\documentclass[journal]{IEEEtran}
\usepackage{cite}
\usepackage{amsmath,amssymb,amsfonts}
\usepackage{graphicx}
\graphicspath{ {Images/} }
\usepackage{textcomp}
\usepackage[table,xcdraw]{xcolor}

\usepackage{hyperref}
\usepackage{array}
\usepackage{multirow}
\usepackage{algorithm}
\usepackage{algpseudocode}
\usepackage{float}

\makeatletter
\newcommand\fs@spaceruled{\def\@fs@cfont{\bfseries}\let\@fs@capt\floatc@ruled
  \def\@fs@pre{\vspace{0.8\baselineskip}\hrule height.8pt depth0pt \kern2pt}%
  \def\@fs@post{\kern2pt\hrule\relax}%
  \def\@fs@mid{\kern2pt\hrule\kern2pt}%
  \let\@fs@iftopcapt\iftrue}
\makeatother

\def\BibTeX{{\rm B\kern-.05em{\sc i\kern-.025em b}\kern-.08em
    T\kern-.1667em\lower.7ex\hbox{E}\kern-.125emX}}

\begin{document}
%
\title{Follow the Gradient: Crossing the Reality Gap using Differentiable Physics (RealityGrad)}
%
%

\author{Jack Collins$^{1,2}$, Ross Brown$^{2}$, J\"urgen Leitner$^{2,3}$ and David Howard$^{1}$%
\thanks{This research was supported by a Data61 PhD Scholarship. J.C acknowledges continued support from the Queensland University of Technology (QUT) through the QUT Centre for Robotics.}
\thanks{$^{1}$ Data61/CSIRO, Brisbane, Australia}%
\thanks{$^{2}$ Queensland University of Technology (QUT), Brisbane, Australia}%
\thanks{$^{3}$ LYRO Robotics Pty Ltd, Brisbane, Australia}%
}

\maketitle

\begin{abstract}

We propose a novel iterative approach for crossing the reality gap that utilises live robot rollouts and differentiable physics.  
Our method, \textit{RealityGrad}, demonstrates for the first time, an efficient sim2real transfer in combination with a real2sim model optimisation for closing the reality gap. 
Differentiable physics has become an alluring alternative to classical rigid-body simulation due to the current culmination of automatic differentiation libraries, compute and non-linear optimisation libraries. Our method builds on this progress and employs differentiable physics for efficient trajectory optimisation.
We demonstrate RealitGrad on a dynamic control task for a serial link robot manipulator and present results that show its efficiency and ability to quickly improve not just the robot's performance in real world tasks but also enhance the simulation model for future tasks.
One iteration of RealityGrad takes less than 22 minutes on a desktop computer while reducing the error by 2/3, making it efficient compared to other sim2real methods in both compute and time. Our methodology and application of differentiable physics establishes a promising approach for crossing the reality gap and has great potential for scaling to complex environments.

\end{abstract}


\IEEEpeerreviewmaketitle

\section{Introduction}

The reality gap is a complex problem in robotics that remains unsolved; it defines the discrepancies experienced when naively transferring a controller developed in simulation to a real world platform \cite{Collins2019QuantifyingTasks}. The gap between simulation and reality is due to the fact that simulation is an inferior replica of the real world and therefore does not capture the noise and stochasticity of the real world or replicate all physical phenomena we experience when interacting with the world around us. With the rise of deep learning and other data-intensive machine learning approaches our reliance on simulation has increased as it provides cheap data for training. This reliance on simulated data has reinforced the reality gap as a critical problem for roboticists.

Physics engines commonly utilised to simulate robotic systems make the assumption that the world is rigid, allowing for a computationally feasible reduction of the real world. Rigid-body physics engines were, until recently, non-differentiable and thus treated as a black box when employing optimisation techniques. With the increasing ubiquity of automatic differentiation libraries, non-linear optimisation libraries and the cost reduction of compute, differentiable physics engines have become an emerging area of research for several disciplines \cite{Degrave2019ARobotics}.

\begin{figure}[t]
\vspace{1mm}
	\centering
	\includegraphics[width=\linewidth]{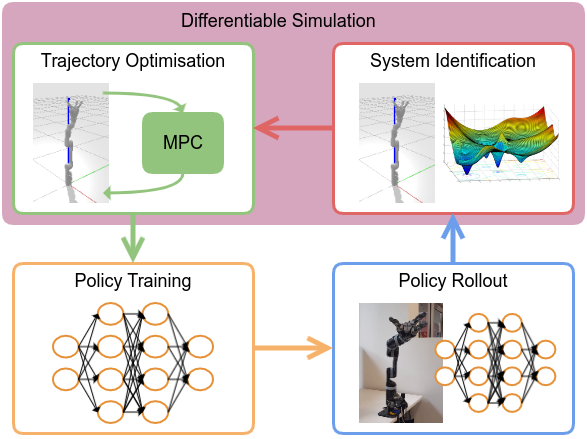}
	\caption{Our RealityGrad pipeline for combining sim2real and real2sim optimisation; (i) generation of $K$ optimal trajectories using optimised parameters (or default simulation properties initially), (ii) policy training using the $K$ optimal trajectories, (iii) policy rollout onto robot and collection of data, (iv) system identification of better simulator parameters to align the model to the real world robot, iterate steps i-iv.}
	\label{flowchart}
	\vspace{-4mm}
\end{figure}

The appeal of differentiable simulators is the ability to use efficient, gradient-based solvers to evaluate gradients between actions, states and model variables. When compared to differentiable data-trained models, differentiable physics engines have an understandable parameterisation of physical properties and the scenes are easily re-configurable. Within robotics, the potential applications for differentiable simulators is large as these simulators promise the gradient based optimisation of simulation parameters, control inputs and agent morphology. Of particular interest is the application of differentiable simulators for crossing the ever-present reality gap.

We present \emph{RealityGrad} (Fig~\ref{flowchart}), a new methodology for crossing the reality gap employing differentiable simulators. Our approach is compute and time efficient whilst also resulting in an improved simulation environment for additional training tasks. The new approach is a demonstration of online sim2real that completes one or more iterations of:
\begin{itemize}
    \item Trajectory Optimisation in simulation (using Model Predictive Control)
    \item Policy Training (a neural network)
    \item Policy Rollout on a real robot
    \item System Identification in simulation
\end{itemize}

There are several contributions in this paper presenting improvements over other existing simulation to real-world (sim2real) transfer methods. 
Our contributions include: (i) proven efficiency in both compute and time gained from using a differentiable simulator for gradient-based optimisation, (ii) an improved simulation environment that can be retained for training of additional, related tasks, and (iii) trained policies with our method are able to learn robust routines that exploit the environment rather than noise-inducing techniques, such as, domain randomisation.

\section{Related Work}

\subsection{Reality Gap}
The reality gap has been identified as an important problem for the robotics community to solve with annual forums and workshops on the subject \cite{Hofer2020PerspectivesWorkshop}. There are competing ideologies on how best to overcome the problem with several prominent methodologies being proposed and researched by different researchers and groups globally. The most common approach is zero-shot or direct transfer which attempts to directly transfer a solution from simulation to the real robot without any intermediary adaptation \cite{Zhao2020Sim-to-RealSurvey}. 

One method for achieving zero-shot transference of policies onto real platforms is Domain Randomisation (DR). DR exposes controllers to a wide range of plausible and implausible scenes during training to make the final policy robust to discrepancies between its training and target environment \cite{Tobin2017DomainWorld}. DR has seen a lot of success in recent years with successes in both computer vision and control policies that were trained solely in simulation working comparably in the real world. However, training an agent using DR can negatively impact training rate and training time with the final policy likely to also favour robustness and conservative actions over exploiting the environment to find the most efficient or fastest policy \cite{Du2021Auto-TunedTransfer}. There are improvements to DR that address some of these issues; Adaptive Domain Randomisation looks to overcome the reduced learning rate by adapting the degree of randomisation of the training environment \cite{OpenAI2019SolvingHand}.

System Identification (systemID) is another zero-shot approach that has been used for crossing the reality gap. In the context of sim2real, systemID utilises real world data to optimise the simulation parameters to closer align the simulator to the real world \cite{Kolev2015PhysicallyContacts}.  The rise in deep learning (DL) has seen DL applied to simulation systemID by a trained network inferring simulation parameters that generate similar real world results \cite{Wu2015Galileo:Learning}. SystemID can be expensive to implement as it typically requires high quality, relevant data from the target platform in the target environment which is costly to collect \cite{Yu2017PreparingIdentification}.

The use of zero-shot sim2real approaches when training deep reinforcement learning (DRL) policies in simulation and successfully transferring these to the real platform are evident in literature. Research that attempts to train a policy to solve complex problems in simulation and successfully transfer the policy across the reality gap oftentimes utilises several sim2real approaches. An example of this is a policy trained in simulation solving a rubiks cube which used both systemID and domain randomisation to transfer the solution but as a result required a large engineering effort and a great amount of compute \cite{OpenAI2019SolvingHand}. 

A competing ideology to zero-shot transfer is online sim2real whereby the system is able to sample real-world data to improve its model of the world throughout training \cite{Zhao2020Sim-to-RealSurvey}. Online sim2real has been employed by the evolutionary robotics community prior to the uptake of deep learning. Koos et al. \cite{Koos2013TheRobotics} used live rollouts of controllers onto the target platform to improve a surrogate model to judge the real-world disparity of simulated controllers. By incorporating both simulated controller evaluations and real-world evaluation of a subset of controllers the researchers were able to ensure that evolution favoured high performing controllers in the real world without having to test every individual.

An example of a DRL approach that utilises relevant experience from the real world is the SimOpt sim2real methodology. Chebotar et al. \cite{Chebotar2019ClosingExperience} used domain randomisation paired with real world rollouts to adapt the parameter distribution and to align the source environment (simulator) to the target environment (real world). This method was successful and enabled the policy to exploit the environment to accomplish the task however the entire process was computationally expensive, requiring 64 GPUs to adapt the policy in the reportedly short time. That amount of compute is unattainable for most researchers to access and adds additional complexity if trying to run online with a robot (i.e. scheduling time on a high performance cluster whilst running a robot).

\subsection{Differentiable Simulators}

Differentiable physics has the promise of solving many of the drawbacks of doing online sim2real with speed and computational efficiency being some of the major supporting points. Degrave et al. \cite{Degrave2019ARobotics} in 2016 proposed a modern simulator for robotics and control that was entirely differentiable. This work highlighted many of the attractions of a differentiable simulator and we have since seen a large uptake of research in this area.

A common demonstration of a task improved by the use of differentiable simulators is identification of parameters. In the past this would require black-box optimisation or calculation of numerical gradients but can now be achieved using gradients calculated using autodiff libraries. Examples of real2sim systemID include the optimisation of a pendulum start state and parameters \cite{Heiden2020AugmentingGap}, soft robot properties \cite{Du2021UnderwaterSimulation} and frictional  properties\cite{Lidec2021DifferentiableIdentification}. However, one of the limiting factors for systemID using a differentiable simulator is the simulation time which has to be on the order of seconds or less.


Another application for differentiable simulators is the task of policy optimisation. There are several ways that differentiable simulators can be utilised to improve agent actions in simulation. Degrave et al. \cite{Degrave2019ARobotics} demonstrated the efficiency of training a policy using back propagation through time for manipulator reaching and quadruped gait optimisation due to the direct gradients a differentiable simulator provides. Heiden et al. \cite{Heiden2019InteractiveSimulation} applied Iterative Linear Quadratic Control, on the linearized dynamics of a cartpole system for direct trajectory optimisation. In addition to applying iLQR to the problem of trajectory optimisation, Heiden et al. used adaptive MPC to successfully demonstrate sim2sim transfer.

Extending upon their earlier work Heiden et al. \cite{Heiden2021NeuralSim:Networks} demonstrated the transference of a quadrupedal walking policy from simulation to a real quadruped using a large number of optimal trajectories generated from simulation. Another sim2real investigation conducted using differentiable simulators is work by Du et al. where the authors implemented a sim2real approach for a underactuated underwater robot \cite{Du2021UnderwaterSimulation}.

There is other notable research into using differentiable simulators that has investigated morphology evolution of robots \cite{Heiden2019InteractiveSimulation} and differentiable physics paired with differentiable rendering \cite{Jatavallabhula2021GradSim:Control}. However there are limitations for differentiable simulators namely that the optimisation landscape is often discontinuous making some optimisations difficult, these often result from contacts, collision and other non-linear effects. Further to this point, not all operations are differentiable making optimal solutions impossible to find from certain states like the optimisation of morphology where addition or subtraction of limbs are required.

We look to extend upon several of these approaches to perform online sim2real by performing system identification and trajectory optimisation to cross the reality gap. The use of differentiable simulators makes our solution compute and time efficient whilst by using an online method we guarantee an improved simulation environment and improved task performance. The iterative process of the online method also makes it likely that the approach will scale to larger problems with more interactions and parameters to tune.

\section{Proposed Approach - RealityGrad}

We present RealityGrad, an iterative, online method for overcoming the reality gap. The methodology is comprised of four steps that when run iteratively one or more times will minimise and overcome the adverse affects of the reality gap (see Algorithm \ref{RG_algo} for details). The steps are:

\begin{itemize}
    \item trajectory optimisation,
    \item policy training,
    \item policy rollout,
    \item and system identification.
\end{itemize}

Once completed, RealityGrad produces an optimised policy that is able to complete the task and in addition a simulation environment that is optimised to the robot and capable of being used as a training environment for further tasks. 

\floatstyle{spaceruled}
\restylefloat{algorithm}
\begin{algorithm}[tb]
\caption{RealityGrad Algorithm for crossing the reality gap}
\label{RG_algo} 
\begin{algorithmic}[1]
\State $\theta$ = HE Initialisation
\State $p_{init}$ =  Default simulator and default robot params
\State $p$ = $p_{init}$
\For{iteration $i \in \{0,...,M\}$ }
    \For{trajectories $t \in \{0,1,...,K\}$}
        \State Randomise $S_{0}$ and $S_{T}$
        \State Run MPC to collect $X$, $U$ using $p$, $S_{0}$ and $S_{T}$
    \EndFor \Comment{lines 5-8: step 1}
    \State Train $\pi_{\theta}$ on K trajectories \Comment{step 2}
    \State Run $\pi_{\theta}$ on platform \Comment{step 3}
    \State System Identification to update $p$ \Comment{step 4}
\EndFor
\end{algorithmic}
\end{algorithm}

RealityGrad is implemented using the Tiny-Differentiable-Simulator (TDS) \cite{Heiden2021NeuralSim:Networks}, one of the few open source differentiable simulators that offers an API, URDF import and exemplar code. TDS is a c++, header-only library capable of employing Automatic Differentiation through its templated format. We integrate TDS with the Adept Automatic Differentiation Library \cite{Hogan2014FastC++} as it is a c++ library capable of forward mode and reverse mode automatic differentiation. A reverse-mode automatic differentiation library is necessary as it greatly improves the running time efficiency for problems where the input dimension is less than the output, although at the expense of RAM due to the recording of operations. We finally integrate Ceres Solver \cite{AgarwalCeresSolver} as the non-linear optimisation library and mlpack \cite{Curtin2018MlpackLibrary} as the machine learning library.

Each of the following subsections discusses in further detail how each step contributes to RealityGrad. 

\subsection{Step 1: Trajectory Optimisation}

The first step within RealityGrad is to collect optimal trajectories in simulation to be used as a training dataset for policy training. Model Predictive Control (MPC) is used to gather a number, $K$, of optimal state-action trajectories ($X$, $U$) of the robot completing the task in simulation. For MPC, the typical dynamic equations of motion used as the system model are replaced by the differentiable simulator. 
The simulation parameters $p$ are improved with each iteration and are initialised with the default parameter $p_{init}$ at the start.

To generate trajectories we leverage the Control Toolbox provided by the ETH Agile and Dexterous Robotics lab~\cite{Giftthaler2018TheControl}. Optimal trajectories are generated using the Gauss-Newton Multiple Shooting algorithm \cite{Neunert2018Whole-BodyQuadrupeds} as it has proven efficiency for whole body control of dynamic systems. Collecting trajectories can be a lengthy step to run if done sequentially but seeing as this is not a computationally expensive task this can be parallelised across available threads. 

Trajectories start states $S_{0}$ and end states $S_T$ are sampled from the same uniform distribution. The MPC setup includes box constraints to limit the allowed joint torques so as to limit the instability of the simulation which can arise from large forces and to reduce the torques executed on the system. Another consideration is the parameterisation of the cost function to ensure realistic and safe trajectories as well as simulation stability.

\subsection{Step 2: Policy Training}

Using the generated state-action pairs $(X, U)$ from the trajectories we regress a feed-forward neural network to predict the required actions $\tau$ to achieve the task given the current state $S_{n}=X_{n}$ and the desired future state $S_{T}$. Training is conducted using the \textit{mlpack} machine learning library. The network architecture comprises of a fully connect feed forward network with two hidden layers of 128 nodes with ReLU activations. The inputs are state position, state velocity and desired positional states with the output being a torque command for each of the controlled joints. Our loss function is the mean squared error between the predicted output and MPC calculated torques. The neural network has its weights initialised using HE initialisation \cite{He2015DelvingClassification} and uses ADAM as the optimiser \cite{Kingma2014Adam:Optimization}. 

Initial experiments omitted this step and ran the MPC directly upon the robot, however for complex systems with long kinematic chains the MPC becomes unstable and is unable to solve for trajectories fast enough to run live. As a result we introduced this step which allows for high frequency control at the expense of training on a larger number of trajectories which take time to compute. 

The first iteration of RealityGrad is used to collect meaningful data on the target platform, that is, data which is relevant to the chosen task. By training a policy and running a policy rollout before the system identification step the user does not require a hand designed controller to collect observations on the platform. With safety constraints like reduced torque bounds and task-space box constraints the policy is capable of running directly on the robot without damaging the robot.


\subsection{Step 3: Policy Rollout}

During the real-world rollout the trained policy executes $\pi_{\theta}$ on the target platform. The policy takes the current state of the robot $S$ as provided by the Robotic Operating System (ROS) and the desired final state $S_{T}$ to infer the required torques $\tau$ for each joint.
\begin{equation}
\label{policy}
\tau = \pi_{\theta}(S, S_{T})
\end{equation}
As the policy is a neural network the policy runs at a higher frequency when compared to MPC paired with a differentiable simulator. When executing the trajectories the state-action pair is recorded.

Safety is an important factor to consider when running trajectories on a real system and as the trajectories are in joint space with torque control there is a high potential for damage to the platform. To limit the risk the rollout is shortened initially, as the requirement for system identification is \textless1sec, and a box constraint added on system states.



\subsection{Step 4: System Identification}

The data recorded from the real world rollout is then used to improve the simulation accuracy. Integrating the TDS, Adept and a non-linear least squares solver within Ceres Solver the simulator parameters $p$ are optimised towards producing the same output results as the real system when given the same commanded action sequence. 

The residual for the optimisation algorithm is calculated using the difference between the real robot and simulated robot compared in joint space, see equation \ref{residual_error}. Where in the first part of equation \ref{residual_error} $n$ is the time step, $j$ is the joint, $q$ is the real joint state and $\hat{q}$ is the simulated joint state. The second part of equation \ref{residual_error} acts as a regularising term for the optimisation, ensuring that the optimisation does not find parameters far away from the measured values. The regularisation term relies on the current parameter $p$, the starting value $p_{init}$ and the proportional gain term $\alpha_{p}$ which is parameter dependant. As the real robot has a sampling rate less than the simulation step size, linear interpolation is used to up-sample the real data so as the two are directly comparable. 

\begin{equation}
\label{residual_error}
residual = \sum_{n}^{N}\sum_{j}^{J}\|q_{n,j}-\hat{q}_{n,j}\|^{2} + \sum_{p}^{P} \alpha_{p} \times (p-p_{init})^{2}
\end{equation}

Due to the dynamic nature of optimising torques for a serial chain the problem requires that the entire process be simulated and optimised as one however for solving kinematic problems without environment interactions the problem could be distributed by each degree of freedom.

This step is computationally the most expensive. Using reverse mode automatic differentiation prevents the need for multiple forward passes of the differentiation software but does require a large amount of RAM to store the tape of operations. As the tape has a maximum size the number of operations has a hard limit with multiple variables effecting its length, i.e. simulation length, simulation step size, number of optimised variables, complexity of simulation. 

The number of optimised parameters does not negatively effect the search time as would be the case if a black-box optimisation algorithm was applied. There are two options for optimising the parameters, one is to use unbounded optimisation and the other is to use parameter bounds. As this is a major design decision we investigate the possibilities further in Section~\ref{Experiments}. Within the bounded or unbounded range of values for each parameter it is likely for there to be a number of local minima and discontinuities. We utilise parallel basin hopping \cite{McCarty2018ParallelOptimization} to overcome the effect of discontinuities on finding a good solution. Parallel basin hopping parallelises the CPU bound search making efficient use of multi-core CPU's, however the limiting factor for parallelising the search is the amount of available RAM to store the tape.

The first iteration of RealityGrad is spawned using the initial simulation parameters $p_{init}$ and the parameters from the URDF of the robot with each additional parallel thread spawned from a uniform distribution with ranges within the minimum and maximum set by the user. Consecutive iterations use the previous iterations optimal values. 

\section{Experiments} \label{Experiments}

Experimentally we begin by comparing the performance of non-linear optimisers used for system identification of simulation parameters to choose the best for application within RealityGrad. We then investigate the ability of RealityGrad to successfully transfer dynamic policies to a manipulator. The configuration for the experiments consists of a Kinova Mico\textsuperscript{2} as the target platform with the TDS simulation environment setup using the official URDF of the Kinova arm. As only simple geometric collision shapes (i.e. pill, plane, sphere ) are available within TDS along with the additional computational overhead and the discontinuities caused by contacts within the optimisation environment we elect to focus on contact-free environments for our initial experiments.

\begin{figure}[tb]
\vspace{2mm}
	\centering
	\includegraphics[width=0.6\linewidth]{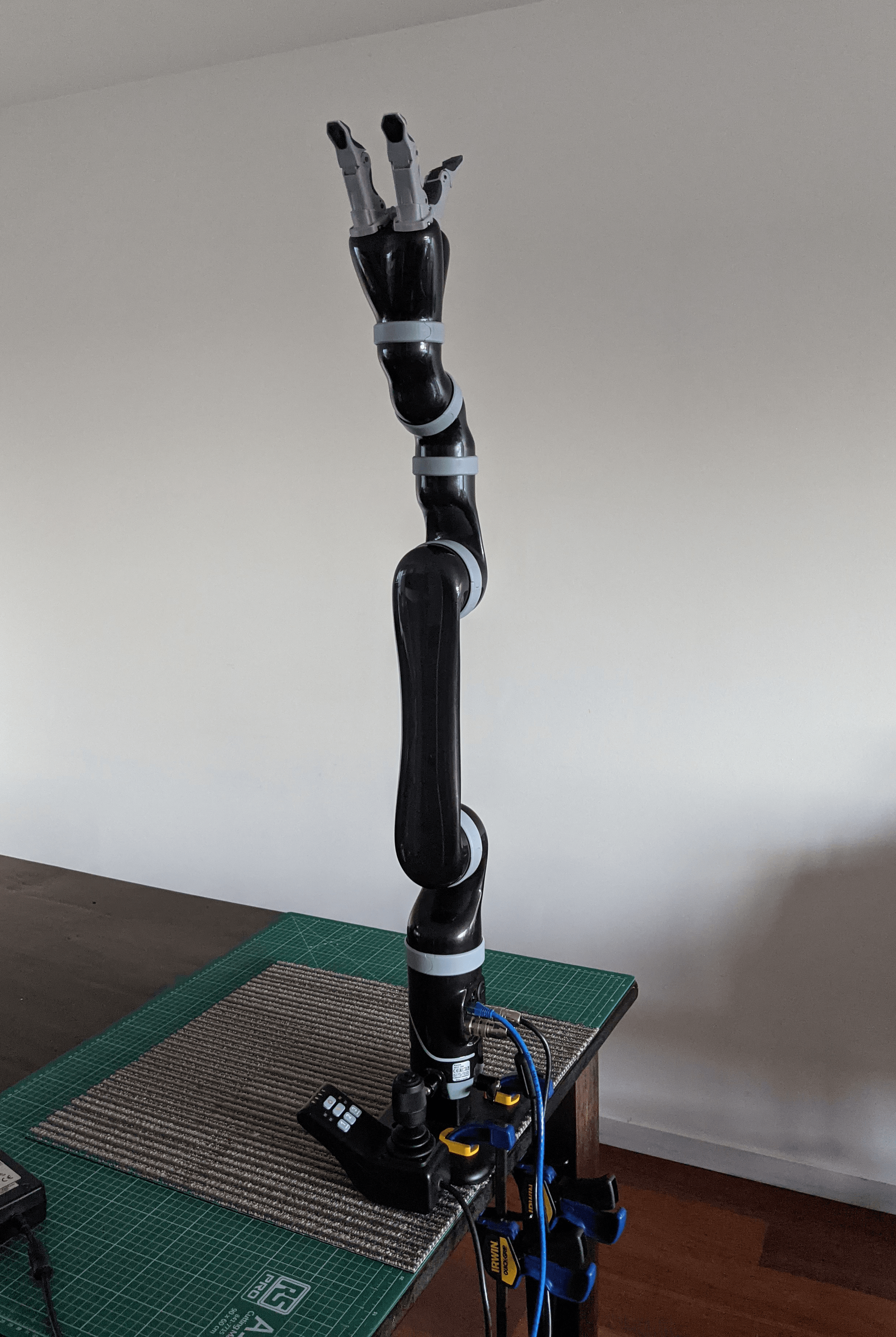}
	\caption{The setup of the Kinova Mico\textsuperscript{2} arm. The ``candle''  pose allows for the zeroing of torque sensors before operation and a safe start configuration to begin torque control. The robot is connected via ethernet connection and with small amendments to the kinova-ros package that inhibits the data published by the publisher to the minimum required for the task so as the communications with the arm is able to achieve 25Hz.}
	\label{robot_setup}
	\vspace{-4mm}
\end{figure}

As the maximum reporting frequency of the Kinova is 25Hz, we set the control frequency to 25Hz for our experiments. We use the kinova-ros package to control the arm and use the reported joint positions and velocities as the state for the policy. For safety the start position for all experiments is the ``candle'' position with joints at $\{0,3.14,3.14,0,0,0\}$ (Figure~\ref{robot_setup}), during torque control the torque safety factor is set to $1.0$ whilst the internal controller monitors for torques that are much larger than gravity compensation values and states that will result in self collisions.

\subsection{Bounded versus Unbounded Non-linear Optimisation}

In this experiment we investigate bounded versus unbounded non-linear optimisation algorithms to inform our decision as to which algorithm we should use within RealityGrad. We optimise the simulation parameters to closer replicate data collected from the real robot. We begin by collecting real world data from the robot by executing a 8second control trajectory on the robot. 

\begin{figure}[tb]
\vspace{2mm}
	\centering
	\includegraphics[width=\linewidth]{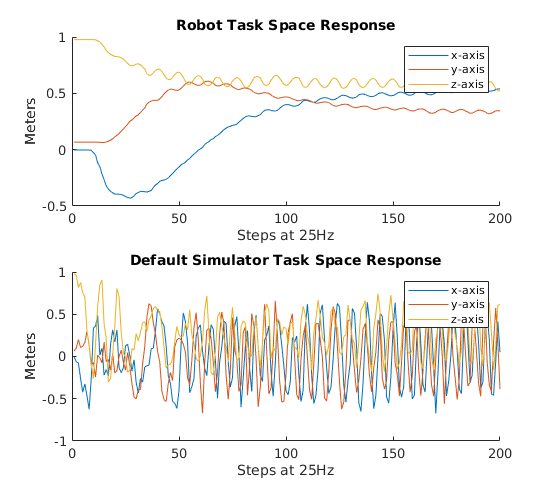}
	\vspace{-4mm}
	\caption{Two plots illustrating the response of the robot in task space (top) and the response of the simulated robot when commanded with the same control input (bottom). The simulated response should look like the robot response however there is a large discrepancy. }
	\label{Robot_vs_sim}
	\vspace{-4mm}
\end{figure}

The trajectory is created using the previously described MPC controller run live with default model/simulation parameters. The randomly chosen desired pose of the end-effector was x: 0.40, y: 0.48, z:0.11. During the real execution of the trajectory the action commands, the joint response and the task-space response were all recorded; the task-space response can be seen in Fig.~\ref{Robot_vs_sim}~(top).

To optimise the simulation the input parameters to the system were improved. There are $71$ parameters we chose to optimise which are summarised below:
\begin{itemize}
    \item gravity, $3$ (x,y,z)
    \item masses, $14$ (each link within the URDF)
    \item damping, $6$ (each controlled joint)
    \item force $P_{d}$, $6$ (proportional factor for control input)
    \item inertias, $3\times14$ (xx, yy, zz for every link)
\end{itemize}

The residual or measure of error between the simulation and real trajectory of the robot was done in joint space and follows Eq. \ref{residual_error} without the second term (regularisation). The optimisation only occurs over a 0.2sec window of the 8sec trajectory due to computational constraints. Within the 0.2sec window the input state is set to the state of the robot, the same control inputs are executed and the simulation is integrated forward in time using the parameters set by the solver.

Ceres solver was the library employed to solve the bounded and unbounded optimisation. The unbounded approach uses the line search limited memory Broyden Fletcher Goldfarb Shanno algorithm \cite{Nocedal1980UpdatingStorage}, whilst the bounded approach uses the trust region Levenberg Marquardt algorithm with bounds. The bounds are $\pm20\%$ for gravity, mass and inertia however for damping and the proportional force parameter we use values of $0\leq damping \leq 18$ and $-1.5\leq P_{d} \leq 1.5$ respectively. Ceres solver is a deterministic solver and therefore multiple runs with the same data find the same solution. The Parallel Basin Hopping optimisation time (60min) and number of parallel workers (6) are held constant for this experiment. 

The simulation of the robot completing the 8sec trajectory in task-space before optimisation of the parameters can be seen in Fig~\ref{Robot_vs_sim}~(bottom). This motivates the need to improve the parameters to ensure the MPC has an accurate model to generate future optimal trajectories and to use for future simulation of additional tasks.


\begin{figure}[tb]
\vspace{2mm}
	\centering
	\includegraphics[width=\linewidth]{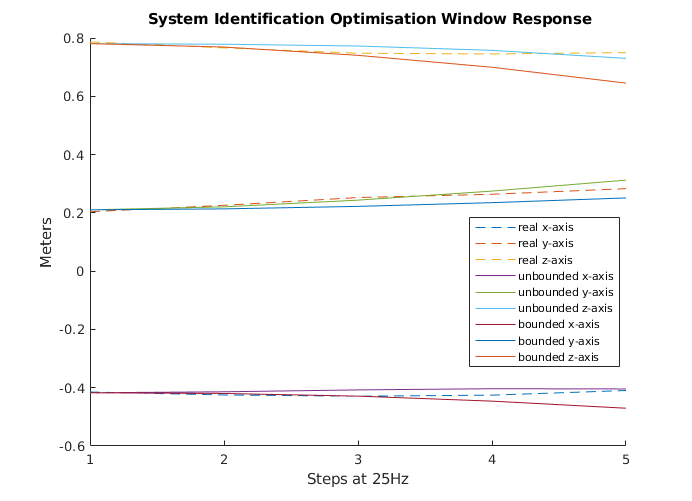}
	\vspace{-4mm}
	\caption{A plot of the optimised response of the unbounded and bounded algorithms in task-space for the 0.2sec optimisation window. The response of the real robot is also plotted as the dashed line. Small deviations in this short optimisation window equate to large deviations over longer simulation times.}
	\label{Short_SI}
	\vspace{-4mm}
\end{figure}

Both the unbounded and bounded optimisation of parameters improved the task-space response of the simulation. The plot in Fig.~\ref{Short_SI} depicts the 0.2sec optimisation window used and the response of the real robot (dashed line) and the simulation when using improved parameters. The residual of the unbounded optimisation achieved a value of $2.6700$ whilst the residual for the bounded optimisation was $130.857$. These values reflect what is seen in Fig.~\ref{Short_SI} as the unbounded approach is better able to mirror the robot response.

\begin{figure}[tb]
\vspace{2mm}
	\centering
	\includegraphics[width=\linewidth]{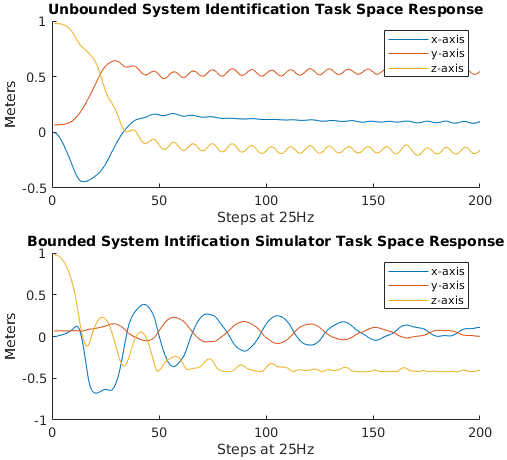}
	\vspace{-4mm}
	\caption{Two plots illustrating the outcome of the unbounded (top) versus bounded (bottom) non-linear optimisation algorithms. Both plots presents the response across a 8sec window after optimising on the same 0.2sec of data. Both plots aim to achieve the response of the robot as seen in Fig. \ref{Robot_vs_sim}~(top). }
	\label{Bounded_vs_Unbounded}
	\vspace{-4mm}
\end{figure}

Extending the simulation from the short 0.2sec window to the entire 8sec simulation of the task-space response we see a much improved trajectory as seen in Fig.~\ref{Bounded_vs_Unbounded}. The cumulative Euclidean error between the real robot and unbounded optimisation across the 8sec was found to be $10.2955m$ whilst the same measure as applied to the bounded optimisation found an error of $14.4414m$. Due to the success of the unbounded approach in both the short window of optimisation and the 8sec extended simulation we elect to use it within further RealityGrad experiments.

Looking further at the data we see that although the unbounded method is able to find parameters that are better for simulating the trajectory it finds parameters that cause the robot to have difficulty opposing gravity (i.e. the z-axis falls to ~-0.1m). Looking at the parameters that likely cause this we see increased weights for each link and changes to the inertial properties. This local minima found by the optimisation is due mainly to the fact that the optimisation occurs over only a 0.2sec period of data, ideally this would be a much larger window. Optimising over a different 0.2sec window of data produces different parameters. Future work will investigate improvements in automatic differentiation and differentiable physics so as longer time sequences can be optimised.

Our intuition on how unbounded optimisation obtained improved results is that although the parameters might not be within feasible real world ranges like those of the bounded optimisation the simulation environment itself is an abstraction leading to abstract parameter values minimising the gap. This, as well as the unimpeded search space that the unbounded algorithm has to work within allows the algorithm to be more expressive and permits the unbounded optimisation the ability to better satisfy the objective.


\subsection{RealityGrad}

This experiment demonstrates the RealityGrad framework by completing two iterations of RealityGrad and analysing the results. The experiment was repeated 5 times with similar results; the results from the best run are discussed below. 

\subsubsection{Step 1} The creation of a dataset of MPC trajectorie begins with 450 trajectories all with randomly initialised starting and end configurations collected. The trajectories all have a length of 1sec. The data collection is parallelised across 30 threads of a desktop with an AMD 3970, taking a total time of 77sec to collect 11,250 data points.

\subsubsection{Step 2} Regress a policy on optimal trajectories collected in step 1 using MPC. Training the policy takes a further 24sec to complete 100 epochs on the 450 trajectories.

\begin{figure*}[ht]
\vspace{2mm}
	\centering
	\includegraphics[width=\linewidth]{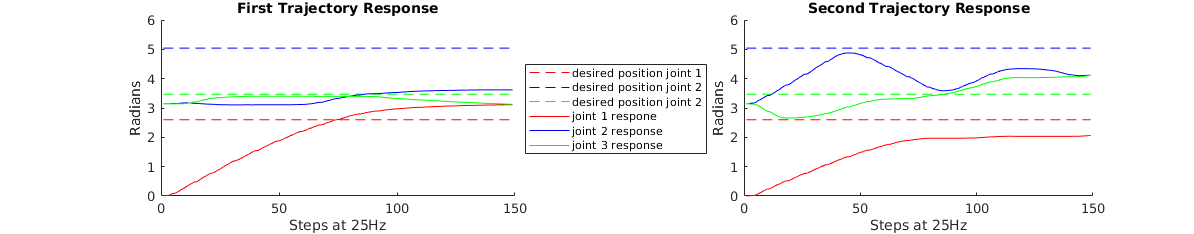}
	\vspace{-4mm}
	\caption{Plot of the two real world executions of the trained policy. The first trajectory (top) is from the first iteration of RealityGrad and the second trajectory (bottom) is from the second iteration of RealityGrad. The plots include the joint response from the first three joints of the manipulator and a dashed line of the desired goal state. The second trajectory shows a much improved response when compared to the first iteration.}
	\label{JointResponse}
	\vspace{-4mm}
\end{figure*}

\subsubsection{Step 3} Executing the policy on the real robot is the next step with a 6sec rollout completed. The graph in Fig.~\ref{JointResponse}~(left) is the plot of the joint response of the first three joints with the dashed lines highlighting the randomly chosen desired pose.

\subsubsection{Step 4} The last step after rolling out the policy on the real robot is system identification. We restrict the amount of time for the non-linear optimisation to only 10mins although the time is only checked at the end of each iteration of the optimiser, making the real optimisation time longer. The actual time for the optimisation was 19.5mins meaning that one iteration of RealityGrad took a total of 21.3mins. 

\begin{figure}[ht]
	\centering
	\includegraphics[width=\linewidth]{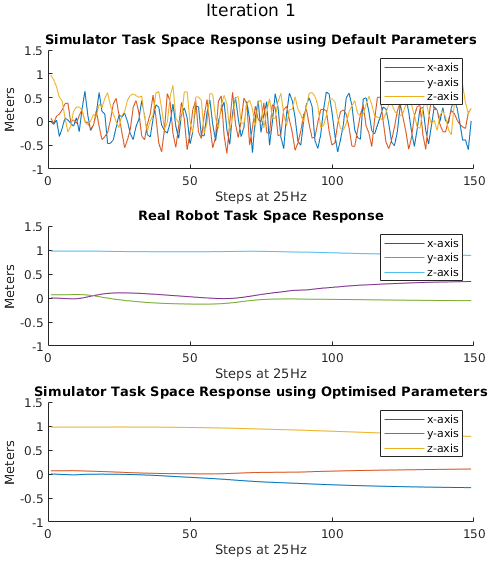}
	\vspace{-4mm}
	\caption{The final step, System Identification, of the first iteration of RealityGrad presented as a subplot in task-space; (top) the real robot response, (middle) the default simulation of the robot trajectory, (bottom) the optimised simulation of the robot trajectory. The optimised simulation (bottom) shows a much closer alignment to the real robot (top).}
	\label{Iteration0}
	\vspace{-4mm}
\end{figure}

The residual for the default parameters before system identification was $350155.24$ with the final optimised residual achieving a value of $28.87$. The accumulated Euclidean error between the trajectory generated using the default simulation and the real robot was $129.79m$ with the accumulated Euclidean error between the optimised simulation trajectory and the real robot being $47.81m$, a vast improvement. The largest difference in parameters was found in the joint damping with values of $\{12.12, 15.18, 17.56, 3.63, 4.31, 4.27\}$. The three plots in Fig. \ref{Iteration0} best illustrate the task-space difference between the response of the simulator with default parameters, the real robot response, and the response of the simulator with optimised parameters.

\subsubsection{Second Iteration - Step 1, 2 and 3} After system identification RealityGrad begins again with new trajectories generated using the optimised simulation environment and a new policy trained on the dataset.

The rollout of the policy trained on the improved simulator trajectories results in much improved joint response as can be seen in Fig.~\ref{JointResponse}~(right). The joint response of the improved model does not converge on the correct position in the second iteration of RealityGrad but instead either oscillates around the correct value or converges on a static position close to the desired position.

\begin{figure}[t]
	\centering
	\includegraphics[width=\linewidth]{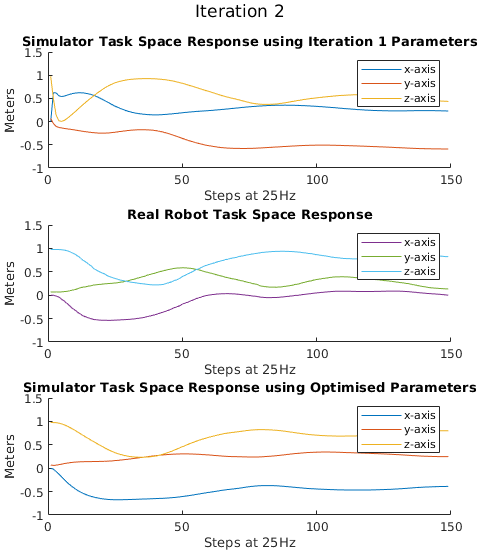}
	\vspace{-4mm}
	\caption{Three subplots of the system identification response in task space for second iteration where: (top) is the simulator response using the optimised parameters from iteration 1, (middle) the real robot response, and (bottom) the simulator response using the new optimised parameters. When comparing }
	\label{Iteration1}
	\vspace{-4mm}
\end{figure}

\subsubsection{Step 4} The second iteration of system identification shows further improvement on the previous iteration's values with the second iteration achieving a residual of $12.26$. Similar to Fig.\ref{Iteration0}, Fig.\ref{Iteration1} presents the system identification response for the second iteration. The new parameters show large improvements over the previous iteration's values with a accumulated Euclidean error of $59.98m$ between the newly optimised simulator parameters and the real robot when compared to $150.72m$ when simulated using the first iteration parameters.

The RealityGrad method can be run further to seek better values for the model but the values found by the second generation are able to accurately predict $>1sec$ into the future which is enough to generate accurate MPC trajectories. Next steps would include training a larger network with many more trajectories to improve generalisation and the response of the joints. 

We see this approach scaling well to problems with increased difficulty by including a curriculum into the task to be executed and choosing the window of time to perform system identification on by measuring the information gain of the system. In addition the total compute was comparatively small with everything being performed on the CPU of a desktop computer (AMD 3970 and 128Gb of RAM), however additional performance could be gained by utilising GPU resources. 


\section{Conclusion}

Differentiable simulators are an exciting avenue of research that have recently become feasible for robotics applications. 
In this paper we present a novel approach that for the first time demonstrate an online sim2real method that uses differentiable physics to efficiently optimise trajectories and at the same time optimises the simulation model for robots in a scalable approach that we title \textit{RealityGrad}. The methodology consists of four steps which include trajectory optimisation (in simulation), policy training (training a neural network to approximate the policy), policy rollout on a real robot and system identification (again in simulation). 
We are able to showcase the efficacy and efficiency when using RealityGrad with articulated robotic manipulators. In our experiments we see a reduction in operational space error when following an arbitrary trajectory by up to 2/3 in a single iteration.

The on-going advancement of differentiable simulators is an exciting avenue to progress sim2real methods. We see potential in upcoming advancements particularly, extending the length and speed of automatic differentiation through the simulator, further improvement in the simulation environments, the augmentation of data models into the physics model of the simulator, and the inclusion of differentiable rendering into crossing the reality gap.







\bibliographystyle{IEEEtran}

\end{document}